\pdfoutput=1

\documentclass[11pt]{article}

\usepackage{acl}

\usepackage{times}
\usepackage{latexsym}
\usepackage{booktabs}
\usepackage{array}
\usepackage{makecell}
\usepackage{amsmath}
\usepackage{graphicx} 
\usepackage{tabularx}
\usepackage{float}
\usepackage{enumitem}

\usepackage[T1]{fontenc}

\usepackage[utf8]{inputenc}

\usepackage{microtype}

\title{An automatically discovered chain-of-thought prompt generalizes to novel models and datasets}

\author{Konstantin Hebenstreit$^{1,2,*}$, Robert Praas$^{1,3,*}$, Louis P Kiesewetter$^{4}$,  Matthias Samwald$^{1,}$\textsuperscript{§} \\
  $^1$Institute of Artificial Intelligence, 
  Medical University of Vienna, Austria \\
  $^2$Johannes Kepler University Linz, Austria \\
  $^3$ KTH Royal Institute of Technology, Stockholm, Sweden \\
  $^4$Humboldt University Berlin, Germany \\
  \textsuperscript{*}equal contribution, 
  \textsuperscript{§}corresponding author \\
  \texttt{matthias.samwald [@] meduniwien.ac.at} \\
  }

\begin{document}
\maketitle
\begin{abstract}
Emergent chain-of-thought (CoT) reasoning capabilities promise to improve performance and explainability of large language models (LLMs). However, uncertainties remain about how reasoning strategies formulated for previous model generations generalize to new model generations and different datasets. In this small-scale study, we compare different reasoning strategies induced by zero-shot prompting across six recently released LLMs (davinci-002, davinci-003, GPT-3.5-turbo, GPT-4, Flan-T5-xxl and Cohere command-xlarge) on a mixture of six question-answering datasets, including datasets from scientific and medical domains.  Our findings demonstrate that while some variations in effectiveness occur, gains from CoT reasoning strategies remain robust across different models and datasets. GPT-4 has the most benefit from current state-of-the-art reasoning strategies and exhibits the best performance by applying a prompt previously discovered through automated discovery.
\end{abstract}

\section{Motivation}

Emergent chain-of-thought (CoT) reasoning capabilities in large language models (LLMs) promise to improve both predictive performance and explainability of models when applied to complex tasks \cite{wei_2021}. While good performance can be reached by few-shot in-context prompting with exemplars suitable to a specific task at hand, zero-shot prompting setups do not require such task-dependent selection of exemplars \cite{kojima_2022}. The recent success of models optimized for dialog, such as GPT-3.5, further increases the expectation that models reach robust performance with ad-hoc reasoning strategies and are less influenced by minor variations.
This study empirically investigates how well previously discovered zero-shot CoT prompting styles generalize to new model generations and datasets and how they compare to newly developed reasoning strategies. We conduct our evaluations on six question-answering datasets of varying levels of complexity, including scientific and medical domains.

\section{Methods}

\subsection{Datasets}
For our study, we used the ThoughtSource framework \cite{ott_other_2023}, which provides a comprehensive meta-dataset and software library designed for streamlined generation, evaluation, and annotation of chain-of-thought (CoT) reasoning. We covered a sizable range of topics and complexity levels by selecting subsamples of six datasets spanning common-sense \cite{talmor_2019, geva-etal-2021-aristotle}, scientific \cite{xie_2020_worldtree, mihaylov_2018_can}, and medical domains \cite{jin_2021_disease, pal_2022_medmcqa} (Table~\ref{table:datasets-descriptions}). All of these question-answering datasets were multiple-choice, consisting of two to five answer options with a single correct response. 

\begin{table*}
\centering
\begin{tabular}{lp{12cm}}
\toprule
\textbf{Dataset} & \textbf{Description} \\
\midrule
CommonsenseQA & General domain crowd-sourced questions with high semantic complexity which command the use of prior knowledge. \\
StrategyQA & General domain crowd-sourced questions which require implicit reasoning and multi-step answer strategies. Yes/No answers. \\
WorldTree v2 & Elementary science questions for 3rd to 5th grade level, combining domain specific and world knowledge. \\
OpenBookQA & Scientific and broad common knowledge questions, which require multi-step reasoning and rich text comprehension. \\
MedQA & Questions from medical board exams. We used only examples from the US (USMLE subset). \\
MedMCQA & Real-world medical entrance exam questions. \\
\bottomrule
\end{tabular}
\caption{\label{table:datasets-descriptions}
Descriptions of various datasets.
}
\end{table*}


\subsection{Prompts}

We assembled a set of ten zero-shot reasoning strategies (Table~\ref{table:prompts-descriptions} and Appendix \ref{appendix:input-templates}) consisting of one baseline, two pre-existing, and seven novel designs:

\begin{enumerate}[itemsep=0.3em]
	\item Direct prompting: No specific trigger, serving as a baseline for comparison.
	\item Kojima: A well-established CoT prompt, "Let's think step by step." \cite{kojima_2022}
	\item Zhou: An enhanced version created through automated prompt engineering, "Let's work this out in a step by step way to be sure we have the right answer." \cite{zhou2023large}
	\item Seven original reasoning strategies designed by us, inspired by various public resources \cite{openai_website_2023,schulhoff_website_2022}, and refined through iterative adaptation based on analyzing outputs. One of these strategies employed a self-critique strategy, requiring the model to provide an initial answer, critique it, and then propose a revised response \cite{madaan_2023,saunders_2022}.
\end{enumerate}

\subsection{Models}

We included six instruction-tuned models based on their reported capabilities in CoT reasoning: davinci-002 \cite{brown_2020}, davinci-003 \cite{ouyang_2022}, GPT-3.5-turbo \cite{openai_website_2022}, and GPT-4 \cite{openai_2023} from OpenAI, Flan-T5-xxl from Google \cite{chung_2022}, and command-xlarge-nightly from Cohere \cite{cohereai_website_2023}. We used the LangChain framework \cite{chase_software_2022} to access several APIs. Between February and April 2023, we conducted 11,880 experiments, all with the model temperature set at 0 and a maximum token length of 512.

\subsection{Evaluation}

We selected Krippendorff's alpha as our evaluation metric \cite{krippendorff_2011}. It allows for combining results from sub-datasets with different numbers of answer choices by correcting for their corresponding base probability rates. Krippendorff's alpha measured inter-rater reliability on a scale from zero (random chance) to one (complete agreement) and was used to compare model predictions to gold standard answers \cite{castro_software_2017}.

To determine an appropriate sample size, we performed a power analysis with a significance level set at 0.05, a medium Krippendorff's alpha value of 0.8, and a base correct probability of 0.2, considering the maximum of five answer options in our sub-datasets (Appendix ~\ref{appendix:power-analysis}). The analysis yielded a required sample size of 164 items, which we increased to 198 items, divided into six sub-datasets of 33 items each.
We used bootstrapping (r~=~1000) to compute means and confidence intervals for the generated results. To guarantee accurate Krippendorff scores, which depend on the number of options, we bootstrapped each sub-dataset individually when needed and calculated confidence intervals by pooling standard deviations.

\section{Results}

All scores within this paper are displayed with 95\% confidence intervals (CI).

\begin{table}[H]
\centering
\begin{tabular}{lcc}
\toprule
          \textbf{Prompt} & \makecell{\textbf{GPT-4 \Large $\alpha$ {\scriptsize(CI)}} \\ \textbf{{\scriptsize n per prompt = 198}}}  & \makecell{\textbf{Model avg. \Large $\alpha$ {\scriptsize(CI)}} \\  \textbf{{\scriptsize n per prompt = 1188}}} \\
\midrule
            Zhou & .83 ({\scriptsize.77, .90}) & .53 ({\scriptsize.50, .57}) \\
          Kojima & .80 ({\scriptsize.73, .87}) & .51 ({\scriptsize.47, .55}) \\
     Zhou-instr. & .79 ({\scriptsize.72, .86}) & .50 ({\scriptsize.46, .54}) \\
      Articulate & .79 ({\scriptsize.71, .86}) & .52 ({\scriptsize.48, .56}) \\
        Rephrase & .78 ({\scriptsize.71, .85}) & .54 ({\scriptsize.51, .58}) \\
            Plan & .77 ({\scriptsize.71, .84}) & .50 ({\scriptsize.46, .54}) \\
       Elaborate & .77 ({\scriptsize.70, .84}) & .51 ({\scriptsize.47, .55}) \\
   Self-critique & .76 ({\scriptsize.69, .84}) & .49 ({\scriptsize.45, .53}) \\
        Converse & .74 ({\scriptsize.66, .81}) & .47 ({\scriptsize.43, .51}) \\
          Direct & .71 ({\scriptsize.64, .79}) & .49 ({\scriptsize.45, .52}) \\

\bottomrule
\end{tabular}

\caption{Krippendorff’s alpha ($\alpha$) performance of prompts averaged over datasets. Average taken solely for GPT-4 and over all six models, n total = 11880.}
\label{table:prompts-krippendorff}
\end{table}

Although the performance of many prompts averaged over all datasets is notably similar, we see that applying reasoning strategies outperforms direct prompting. A closer examination of the results obtained from the latest model, GPT-4, highlights the advantage of employing specific prompts (Table~\ref{table:prompts-krippendorff}). It shows the retained performance of the automatically discovered prompt by \citet{zhou2023large}, which also has a notable result in the score averaged over models. Interestingly, the self-critique prompt yielded relatively low scores. It also resulted in the generation of multiple answers in various observed instances, which were excluded from the scoring process. Creating an instruction prompt by placing Zhou’s reasoning prompt before instead of after the question did not yield better outcomes.

Better models are finding WorldTree v2 and CommonsenseQA increasingly easy (Table~\ref{table:datasets-models-krippendorff}), while StrategyQA suffers from peculiar items. This highlights the necessity for developing more refined general-knowledge datasets or employing domain-specific datasets, such as the two medical ones (Table~\ref{table:datasets-krippendorff}).

\begin{table}[H]
\centering
\begin{tabular}{lcc}
\toprule
        \textbf{Dataset} &       \makecell{\textbf{ \Large $\alpha$ {\scriptsize(CI)}} \\  \textbf{{\scriptsize n per dataset = 1980}}} \\
\midrule
 WorldTree v2 & .83 ({\scriptsize.81, .85}) \\
CommonsenseQA & .71 ({\scriptsize.68, .73}) \\
   OpenBookQA & .65 ({\scriptsize.63, .68}) \\
   StrategyQA & .31 ({\scriptsize.27, .36}) \\
      MedMCQA & .31 ({\scriptsize.28, .34}) \\
        MedQA & .21 ({\scriptsize.19, .24}) \\
\bottomrule
\end{tabular}

\caption{Krippendorff’s alpha ($\alpha$) performance on data-sets averaged over models and prompts. N total = 11880. }
\label{table:datasets-krippendorff}
\end{table}

FLAN-T5 shows good performance for its size, but its results are possibly affected by data contamination: It was instruction-fine tuned on the sub-datasets CommonsenseQA and StrategyQA. GPT-3.5-turbo and GPT-4 performed best (Table~\ref{table:models-krippendorff}) and were the only models that displayed decent performance on medical datasets (Table~\ref{table:datasets-models-krippendorff}).

\begin{table}[H]
\centering
\begin{tabular}{lcc}
\toprule
        \textbf{Model} &       \makecell{\textbf{ \Large $\alpha$ {\scriptsize(CI)}} \\  \textbf{{\scriptsize n per model = 1980}}} \\
\midrule
        GPT-4 & .78 ({\scriptsize.76, .81}) \\
GPT-3.5-turbo & .62 ({\scriptsize.59, .65}) \\
  Davinci-003 & .47 ({\scriptsize.45, .50}) \\
  Flan-T5-XXL & .45 ({\scriptsize.42, .47}) \\
  Davinci-002 & .41 ({\scriptsize.38, .44}) \\
   Command-XL & .32 ({\scriptsize.29, .35}) \\
\bottomrule
\end{tabular}

\caption{Krippendorff’s alpha ($\alpha$) performance of models averaged over datasets and prompts. N total = 11880. }
\label{table:models-krippendorff}
\end{table}

Further detailed results, as well as results reported as accuracy values can be found in the appendix.

\section{Limitations}

The presented work has several limitations. Our study aimed to test a wide variety of combinations of prompts, datasets, and models under budgetary constraints. We therefore chose to subsample datasets based on a statistical power analysis. This limits the direct comparison of our results to evaluations on full benchmark test sets.
Upon inspecting results for some of the academic benchmark datasets generated through crowdsourcing we found that the quality of a sizable subset of examples was not optimal. One common pattern we found was that questions and answer choices did not allow for clearly picking a best answer. More advanced models tend to correctly point out such problems in their reasoning response and refrain from selecting a single answer choice.
We did not use methods such as self-consistency \cite{wang_2022c} that maximize final accuracy at the expense of practical interpretability, i.e., we targeted situations in which users expect a single, high-quality and easily interpretable reasoning chain rather than a collection of noisy reasoning chains. Results achieved when using prompts in conjunction with ensemble methods might potentially differ.
Our study included state-of-the-art closed-source models which are undergoing constant change, making replication and comparisons over time difficult. We partially address this concern by making all data generated by models at the time of our experiment openly available.
The lack of documentation of closed models also leads to concerns about contamination of training data with benchmark datasets. While our comparison of different prompts is not severely impacted, we caution against strongly interpreting results across different models for this reason. We noted that Flan-T5 \cite{longpre_2023}, which was instruction-finetuned on the subsets of CommonsenseQA and StrategyQA, outperformed GPT-3.5-turbo on CommonsenseQA.

\section{Discussion}

\textbf{Related work.} Several related studies evaluated zero-shot prompting performance. As a notable example, Liévin \cite{livin_2022} performed a comparable zero-shot CoT evaluation focused on medical datasets. Earlier work evaluating multiple models and datasets zero-shot includes commonsense data \cite{zhou_2020a} and the assessment of T0 performance on multiple-choice tasks \cite{orlanski_2022}. HELM \cite{liang_2022} covers a wide range of model comparisons. Our study added to current knowledge by focusing on finding simple and versatile chain-of-thought prompting approaches that work across a spectrum of question-answering datasets and models.

\textbf{Future work.} The current study can be extended by evaluating prompts and datasets with additional models, particularly the multitude of openly available LLMs like LLaMa, the Pythia suite, dialog-tuned models like Alpaca  \cite{touvron_2023,biderman_2023,taori_2023}, StableLM \cite{stability_web}, and OpenAssistant \cite{laion}. Finally, user evaluations of the quality and explanatory utility of reasoning chains generated by different prompts and models need to be conducted.

\section{Acknowledgements}

We thank the Cohere team for providing custom API access, enabling faster inference and unrestricted analysis of medical question datasets that were occasionally flagged by the standard API.

\bibliography{anthology,custom,sci_wheel}
\bibliographystyle{acl_natbib}

\appendix

\newpage
\section{Model input templates}
\label{appendix:input-templates}

This Appendix provides an overview of the text templates and prompt structures used in our research. The model input is structured as follows:\\

\begin{verbatim}
{instruction}

{question}
{answer_choices}

{cot_trigger}
\end{verbatim}
\vspace{5mm}
Placeholder descriptions:
\begin{itemize}
    \item \{question\}: The multiple-choice question that the model is expected to answer.
    \item \{answer\_choices\}: The options provided for the multiple-choice question. 
\end{itemize}
Two methods of prompting are employed (only one is used at a time):
\begin{itemize}
    \item \{instruction\}: Placed before the question and answer choices.
    \item \{cot\_trigger\}: Placed after the question.
\end{itemize}
Each prompt is accompanied by its type in parentheses in Table~\ref{table:prompts-descriptions}.

\section{Power Calculation Formula}
\label{appendix:power-analysis}

\[T\left(P_c, \alpha_{\text {min}}, p\right)=\]

\[=\\2 z_p^2\left(\frac{\left(1+\alpha_{\text {min}}\right)\left(3-\alpha_{\text {min}}\right)}{4\left(1-\alpha_{\text {min}}\right) P_c\left(1-P_c\right)}-\alpha_{\text {min}}\right)\]

Where:
\begin{align*}
P_c &= \text{the probability of value } c \\
\alpha_\text{min} &= \text{the smallest } \alpha \text{ for coding to be accepted } \\
&\phantom{=} \text{ as reliable} \\
p &= \text{level of significance} \\
z_p &= \text{the standardized } z \text{-statistics at } p
\end{align*}

\section{Tables}

Some prompts work specifically well on certain datasets. The rephrasing prompt seems to help with the ambiguous questions which we found to be prevalent in the StrategyQA dataset.

\begin{table*}
\centering
\renewcommand{\arraystretch}{1.2} 
\begin{tabular}{p{3cm}p{3cm}p{8cm}}
\toprule
\textbf{ID} & \textbf{Prompt Name} & \textbf{Text} \\
\midrule
None & \textit{Direct} & "{Direct prompting. No specific prompt is used. Just the question and answer choices are the input to the model.}" \\
kojima-01 (cot\_trigger) & \textit{Kojima} & "Answer: Let's think step by step." \\
zhou-01 (cot\_trigger) & \textit{Zhou} & "Answer: Let's work this out in a step by step way to be sure we have the right answer." \\
\raisebox{-2ex}{zhou-01-ins}\\(instruction) & \textit{Zhou-instruction} & "Let's work this out in a step by step way to be sure we have the right answer." \\
qa-10 (instruction) & \textit{Plan} & "First think step by step - describe your plan for how to get to the right answer, written out in great detail. Then answer the question." \\
qa-12 (instruction) & \textit{Articulate} & "Carefully read the question \& work this out in a step by step way to be sure you have the right answer. Be certain to spell out your thoughts \& reasoning so anyone can verify them. Spell out everything in painstaking detail \& don't skip any steps!" \\
qa-13 (instruction) & \textit{Rephrase} & "Instruction: First let's rephrase the question to be sure we understood it correctly. Second, let's work this out step by step by spelling out our thoughts \& reasoning so anyone can verify them. Third, make sure we have the right answer." \\
qa-16 (instruction) & \textit{Elaborate} & "Answer the following question through careful, concise step-by-step reasoning. First, complement the question with helpful knowledge and important additional facts. Second, generate sub-questions that are required to answer the original question, answer them until you can answer the original question." \\
qa-17 (instruction) & \textit{Converse} & "Create a dialog between a professor and a student. The student asks sub-questions to the question. The professor works them out in a step by step way and makes sure that the student understood how they got to the right answer." \\
refl-01 (instruction) & \textit{Self-critique} & "Answer the question, then critique the answer. Based on the critique, reconsider the other answer options and give a single final answer." \\
\bottomrule
\end{tabular}
\caption{\label{table:prompts-descriptions}
Used prompts and their corresponding ID and text.
}
\end{table*}

\begin{table*}
\centering
\begin{tabular}{lcc}
\toprule
          \textbf{Prompt} & \makecell{\textbf{Model avg. accuracy {\scriptsize(CI)}} \\  \textbf{{\scriptsize n per prompt = 1188}}} \\
\midrule
                Zhou & .68 ({\scriptsize.65, .70}) \\
          Articulate & .67 ({\scriptsize.64, .70}) \\
            Rephrase & .67 ({\scriptsize.64, .69}) \\
           Elaborate & .66 ({\scriptsize.63, .69}) \\
    Zhou-instruction & .65 ({\scriptsize.63, .68}) \\
                Plan & .65 ({\scriptsize.62, .68}) \\
              Kojima & .64 ({\scriptsize.62, .67}) \\
              Direct & .64 ({\scriptsize.61, .67}) \\
       Self-critique & .64 ({\scriptsize.61, .67}) \\
            Converse & .64 ({\scriptsize.61, .66}) \\

\bottomrule
\end{tabular}

\caption{Accuracy of prompts averaged over datasets. In Table~\ref{table:prompts-descriptions}, text corresponding to the prompt names can be found. Average taken over all six models. N total = 11880.}
\label{tab:your_label}
\end{table*}

\begin{table*}
\centering
\begin{tabular}{lcc}
\toprule
        \textbf{Dataset} & \makecell{\textbf{ Accuracy {\scriptsize(CI)}} \\  \textbf{{\scriptsize n per dataset = 1980}}} & \textbf{Base Rate} \\
\midrule
         WorldTree v2 & .88 ({\scriptsize.86, .89}) & .25 \\
        CommonsenseQA & .77 ({\scriptsize.75, .79}) & .2 \\
           OpenBookQA & .74 ({\scriptsize.72, .76}) & .25 \\
           StrategyQA & .67 ({\scriptsize.65, .69}) & .5 \\
              MedMCQA & .49 ({\scriptsize.46, .51}) & .25 \\
                MedQA & .38 ({\scriptsize.36, .40}) & .2 \\
\bottomrule
\end{tabular}

\caption{Accuracy on datasets averaged over models and prompts. Base rate for random chance, dependent on number of answer choices in datasets. N total = 11880} 
\label{tab:your_label}
\end{table*}

\begin{table*}
\centering
\begin{tabular}{lcc}
\toprule
        \textbf{Model} &       \makecell{\textbf{ Accuracy {\scriptsize(CI)}} \\  \textbf{{\scriptsize n per model = 1980}}} \\
\midrule
                GPT-4 & .85 ({\scriptsize.83, .86}) \\
        GPT-3.5-turbo & .74 ({\scriptsize.72, .76}) \\
          Davinci-003 & .63 ({\scriptsize.61, .65}) \\
          Flan-T5-XXL & .61 ({\scriptsize.59, .63}) \\
          Davinci-002 & .59 ({\scriptsize.56, .61}) \\
           Command-XL & .52 ({\scriptsize.50, .55}) \\
\bottomrule
\end{tabular}

\caption{Accuracy of models averaged over datasets and prompts. N total = 11880. }
\label{tab:your_label}
\end{table*}

\newpage
\clearpage

\begin{table*}
    \centering
    \resizebox{\textwidth}{!}{%
        \begin{tabular}{lcccccc}
        \toprule
        model &                    Command-XL &                   Flan-T5-XXL &                 GPT-3.5-turbo &                         GPT-4 &                   Davinci-002 &                   Davinci-003 \\
        dataset       &                               &                               &                               &                               &                               &                               \\
        \midrule
        CommonsenseQA &  .57 ({\scriptsize .50, .64}) &  .81 ({\scriptsize .75, .85}) &  .70 ({\scriptsize .64, .76}) &  .82 ({\scriptsize .76, .87}) &  .68 ({\scriptsize .62, .74}) &  .68 ({\scriptsize .62, .74}) \\
        MedQA         &  .06 ({\scriptsize .01, .13}) &  .02 ({\scriptsize .00, .07}) &  .40 ({\scriptsize .32, .47}) &  .55 ({\scriptsize .47, .61}) &  .09 ({\scriptsize .03, .15}) &  .17 ({\scriptsize .11, .24}) \\
        MedMCQA       &  .08 ({\scriptsize .01, .14}) &  .10 ({\scriptsize .03, .17}) &  .51 ({\scriptsize .44, .58}) &  .73 ({\scriptsize .67, .79}) &  .20 ({\scriptsize .13, .27}) &  .21 ({\scriptsize .14, .28}) \\
        OpenBookQA    &  .43 ({\scriptsize .36, .50}) &  .69 ({\scriptsize .63, .76}) &  .77 ({\scriptsize .71, .83}) &  .91 ({\scriptsize .87, .95}) &  .45 ({\scriptsize .37, .52}) &  .66 ({\scriptsize .59, .72}) \\
        StrategyQA    &  .10 ({\scriptsize .00, .21}) &  .23 ({\scriptsize .12, .34}) &  .44 ({\scriptsize .33, .55}) &  .69 ({\scriptsize .61, .76}) &  .20 ({\scriptsize .09, .32}) &  .22 ({\scriptsize .12, .31}) \\
        WorldTree v2  &  .67 ({\scriptsize .61, .73}) &  .77 ({\scriptsize .72, .83}) &  .89 ({\scriptsize .85, .93}) &  .97 ({\scriptsize .95, .99}) &  .84 ({\scriptsize .79, .89}) &  .84 ({\scriptsize .80, .89}) \\
        \bottomrule
        \end{tabular}
    }
        \caption{Krippendorff’s alpha ($\alpha$) performance of models per dataset averaged over prompts. The low score of Flan-T5-XXL on MedQA illustrates that Krippendorff’s alpha ($\alpha$) corrects the accuracy in Table 10 for the base rate in Table 7.}
\label{table:datasets-models-krippendorff}
\end{table*}

\begin{table*}
    \centering
    \resizebox{\textwidth}{!}{%
        \begin{tabular}{lcccccc}
        \toprule
        model &                    Command-XL &                   Flan-T5-XXL &                 GPT-3.5-turbo &                         GPT-4 &                   Davinci-002 &                   Davinci-003 \\
        dataset       &                               &                               &                               &                               &                               &                               \\
        \midrule
        CommonsenseQA &  .66 ({\scriptsize .61, .71}) &  .85 ({\scriptsize .81, .89}) &  .76 ({\scriptsize .71, .81}) &  .85 ({\scriptsize .81, .90}) &  .75 ({\scriptsize .70, .79}) &  .75 ({\scriptsize .70, .80}) \\
        MedQA         &  .27 ({\scriptsize .22, .32}) &  .22 ({\scriptsize .17, .26}) &  .53 ({\scriptsize .47, .58}) &  .65 ({\scriptsize .60, .70}) &  .28 ({\scriptsize .23, .33}) &  .35 ({\scriptsize .30, .40}) \\
        MedMCQA       &  .31 ({\scriptsize .26, .36}) &  .35 ({\scriptsize .30, .40}) &  .63 ({\scriptsize .58, .69}) &  .80 ({\scriptsize .76, .85}) &  .41 ({\scriptsize .35, .46}) &  .41 ({\scriptsize .36, .47}) \\
        OpenBookQA    &  .58 ({\scriptsize .52, .63}) &  .78 ({\scriptsize .73, .82}) &  .83 ({\scriptsize .79, .88}) &  .93 ({\scriptsize .91, .96}) &  .59 ({\scriptsize .54, .65}) &  .75 ({\scriptsize .70, .80}) \\
        StrategyQA    &  .57 ({\scriptsize .51, .62}) &  .62 ({\scriptsize .57, .68}) &  .73 ({\scriptsize .68, .79}) &  .85 ({\scriptsize .81, .89}) &  .63 ({\scriptsize .57, .68}) &  .63 ({\scriptsize .58, .69}) \\
        WorldTree v2  &  .75 ({\scriptsize .71, .80}) &  .83 ({\scriptsize .79, .87}) &  .92 ({\scriptsize .89, .95}) &  .98 ({\scriptsize .96, .99}) &  .88 ({\scriptsize .85, .92}) &  .88 ({\scriptsize .85, .92}) \\
        \bottomrule
        \end{tabular}
    }
        \caption{Accuracy of models per dataset averaged over prompts.}
\label{your-label}
\end{table*}

\begin{table*}
    \centering
    \resizebox{\textwidth}{!}{%
        \begin{tabular}{lcccccc}
        \toprule
        dataset &                 CommonsenseQA &                         MedQA &                       MedMCQA &                    OpenBookQA &                    StrategyQA &                  WorldTree v2 \\
        prompt           &                               &                               &                               &                               &                               &                               \\
        \midrule
        Direct           &  .68 ({\scriptsize .60, .76}) &  .21 ({\scriptsize .12, .30}) &  .28 ({\scriptsize .18, .37}) &  .65 ({\scriptsize .56, .73}) &  .24 ({\scriptsize .10, .38}) &  .84 ({\scriptsize .77, .90}) \\
        Kojima           &  .69 ({\scriptsize .61, .77}) &  .22 ({\scriptsize .14, .31}) &  .25 ({\scriptsize .16, .35}) &  .61 ({\scriptsize .52, .70}) &  .46 ({\scriptsize .32, .59}) &  .79 ({\scriptsize .72, .86}) \\
        Zhou             &  .72 ({\scriptsize .64, .79}) &  .23 ({\scriptsize .14, .32}) &  .37 ({\scriptsize .27, .46}) &  .74 ({\scriptsize .66, .81}) &  .32 ({\scriptsize .19, .44}) &  .83 ({\scriptsize .77, .89}) \\
        Plan             &  .73 ({\scriptsize .65, .80}) &  .19 ({\scriptsize .11, .28}) &  .30 ({\scriptsize .21, .40}) &  .65 ({\scriptsize .56, .73}) &  .27 ({\scriptsize .12, .42}) &  .82 ({\scriptsize .75, .88}) \\
        Articulate       &  .72 ({\scriptsize .64, .80}) &  .22 ({\scriptsize .14, .31}) &  .35 ({\scriptsize .25, .45}) &  .67 ({\scriptsize .59, .75}) &  .27 ({\scriptsize .13, .40}) &  .88 ({\scriptsize .83, .93}) \\
        Rephrase         &  .75 ({\scriptsize .68, .82}) &  .21 ({\scriptsize .13, .29}) &  .31 ({\scriptsize .22, .41}) &  .61 ({\scriptsize .51, .70}) &  .42 ({\scriptsize .30, .55}) &  .87 ({\scriptsize .82, .92}) \\
        Elaborate        &  .68 ({\scriptsize .60, .76}) &  .25 ({\scriptsize .17, .34}) &  .36 ({\scriptsize .25, .45}) &  .64 ({\scriptsize .56, .72}) &  .33 ({\scriptsize .20, .47}) &  .82 ({\scriptsize .75, .88}) \\
        Converse         &  .63 ({\scriptsize .55, .72}) &  .20 ({\scriptsize .12, .29}) &  .32 ({\scriptsize .23, .41}) &  .63 ({\scriptsize .55, .72}) &  .30 ({\scriptsize .16, .43}) &  .78 ({\scriptsize .71, .85}) \\
        Self-critique    &  .73 ({\scriptsize .65, .80}) &  .19 ({\scriptsize .11, .27}) &  .25 ({\scriptsize .16, .34}) &  .66 ({\scriptsize .56, .74}) &  .23 ({\scriptsize .09, .37}) &  .82 ({\scriptsize .75, .88}) \\
        Zhou-instruction &  .73 ({\scriptsize .66, .81}) &  .19 ({\scriptsize .11, .28}) &  .26 ({\scriptsize .17, .36}) &  .65 ({\scriptsize .57, .74}) &  .28 ({\scriptsize .14, .42}) &  .86 ({\scriptsize .80, .92}) \\
        \bottomrule
        \end{tabular}
    }
        \caption{Krippendorff’s alpha ($\alpha$) performance of prompts per dataset averaged over models.}
\label{your-label}
\end{table*}

\begin{table*}
    \centering
    \resizebox{\textwidth}{!}{%
        \begin{tabular}{lcccccc}
        \toprule
        dataset &                 CommonsenseQA &                         MedQA &                       MedMCQA &                    OpenBookQA &                    StrategyQA &                  WorldTree v2 \\
        prompt           &                               &                               &                               &                               &                               &                               \\
        \midrule
        Direct           &  .74 ({\scriptsize .68, .81}) &  .38 ({\scriptsize .31, .45}) &  .46 ({\scriptsize .39, .53}) &  .74 ({\scriptsize .68, .80}) &  .64 ({\scriptsize .57, .71}) &  .88 ({\scriptsize .84, .93}) \\
        Kojima           &  .75 ({\scriptsize .69, .81}) &  .39 ({\scriptsize .32, .46}) &  .44 ({\scriptsize .37, .51}) &  .71 ({\scriptsize .65, .78}) &  .73 ({\scriptsize .67, .80}) &  .85 ({\scriptsize .79, .90}) \\
        Zhou             &  .78 ({\scriptsize .72, .84}) &  .40 ({\scriptsize .33, .47}) &  .54 ({\scriptsize .47, .61}) &  .81 ({\scriptsize .75, .87}) &  .67 ({\scriptsize .60, .74}) &  .87 ({\scriptsize .83, .92}) \\
        Plan             &  .78 ({\scriptsize .73, .84}) &  .36 ({\scriptsize .30, .43}) &  .48 ({\scriptsize .41, .56}) &  .74 ({\scriptsize .68, .80}) &  .66 ({\scriptsize .59, .73}) &  .87 ({\scriptsize .82, .91}) \\
        Articulate       &  .78 ({\scriptsize .72, .84}) &  .39 ({\scriptsize .32, .46}) &  .52 ({\scriptsize .44, .59}) &  .76 ({\scriptsize .70, .82}) &  .66 ({\scriptsize .59, .73}) &  .91 ({\scriptsize .87, .95}) \\
        Rephrase         &  .80 ({\scriptsize .75, .86}) &  .38 ({\scriptsize .31, .45}) &  .49 ({\scriptsize .42, .56}) &  .71 ({\scriptsize .64, .77}) &  .72 ({\scriptsize .65, .78}) &  .91 ({\scriptsize .87, .95}) \\
        Elaborate        &  .75 ({\scriptsize .68, .81}) &  .41 ({\scriptsize .34, .48}) &  .53 ({\scriptsize .46, .60}) &  .74 ({\scriptsize .67, .80}) &  .68 ({\scriptsize .61, .75}) &  .87 ({\scriptsize .82, .92}) \\
        Converse         &  .71 ({\scriptsize .64, .78}) &  .37 ({\scriptsize .30, .44}) &  .51 ({\scriptsize .43, .58}) &  .73 ({\scriptsize .67, .79}) &  .66 ({\scriptsize .59, .73}) &  .84 ({\scriptsize .79, .89}) \\
        Self-critique    &  .79 ({\scriptsize .73, .84}) &  .37 ({\scriptsize .30, .43}) &  .44 ({\scriptsize .37, .51}) &  .75 ({\scriptsize .69, .81}) &  .62 ({\scriptsize .55, .69}) &  .87 ({\scriptsize .82, .92}) \\
        Zhou-instruction &  .79 ({\scriptsize .73, .85}) &  .37 ({\scriptsize .30, .44}) &  .45 ({\scriptsize .38, .52}) &  .74 ({\scriptsize .68, .81}) &  .66 ({\scriptsize .59, .73}) &  .90 ({\scriptsize .86, .94}) \\
        \bottomrule
        \end{tabular}
    }
        \caption{Accuracy of prompts per dataset averaged over models.}
\label{your-label}
\end{table*}

\clearpage

\begin{table*}
    \centering
    \resizebox{\textwidth}{!}{%
        \begin{tabular}{lcccccc}
        \toprule
        model &                    Command-XL &                   Flan-T5-XXL &                 GPT-3.5-turbo &                         GPT-4 &                   Davinci-002 &                   Davinci-003 \\
        prompt           &                               &                               &                               &                               &                               &                               \\
        \midrule
        Direct           &  .26 ({\scriptsize .18, .33}) &  .49 ({\scriptsize .41, .58}) &  .61 ({\scriptsize .53, .69}) &  .71 ({\scriptsize .64, .79}) &  .41 ({\scriptsize .31, .50}) &  .44 ({\scriptsize .35, .53}) \\
        Kojima           &  .25 ({\scriptsize .16, .34}) &  .46 ({\scriptsize .38, .55}) &  .66 ({\scriptsize .57, .75}) &  .80 ({\scriptsize .73, .87}) &  .42 ({\scriptsize .33, .51}) &  .45 ({\scriptsize .36, .54}) \\
        Zhou             &  .35 ({\scriptsize .27, .43}) &  .44 ({\scriptsize .37, .51}) &  .62 ({\scriptsize .53, .71}) &  .83 ({\scriptsize .77, .90}) &  .53 ({\scriptsize .45, .62}) &  .50 ({\scriptsize .41, .59}) \\
        Plan             &  .34 ({\scriptsize .25, .42}) &  .45 ({\scriptsize .37, .53}) &  .61 ({\scriptsize .52, .70}) &  .77 ({\scriptsize .71, .84}) &  .37 ({\scriptsize .30, .45}) &  .46 ({\scriptsize .37, .55}) \\
        Articulate       &  .33 ({\scriptsize .26, .40}) &  .50 ({\scriptsize .42, .58}) &  .59 ({\scriptsize .49, .68}) &  .79 ({\scriptsize .71, .86}) &  .44 ({\scriptsize .35, .53}) &  .52 ({\scriptsize .43, .60}) \\
        Rephrase         &  .42 ({\scriptsize .33, .51}) &  .46 ({\scriptsize .38, .54}) &  .61 ({\scriptsize .52, .70}) &  .78 ({\scriptsize .71, .85}) &  .44 ({\scriptsize .35, .53}) &  .46 ({\scriptsize .37, .55}) \\
        Elaborate        &  .34 ({\scriptsize .26, .42}) &  .42 ({\scriptsize .33, .51}) &  .61 ({\scriptsize .51, .70}) &  .77 ({\scriptsize .70, .84}) &  .51 ({\scriptsize .42, .60}) &  .43 ({\scriptsize .35, .51}) \\
        Converse         &  .31 ({\scriptsize .22, .40}) &  .44 ({\scriptsize .35, .52}) &  .58 ({\scriptsize .49, .67}) &  .74 ({\scriptsize .66, .81}) &  .35 ({\scriptsize .26, .43}) &  .46 ({\scriptsize .38, .54}) \\
        Self-critique    &  .32 ({\scriptsize .26, .39}) &  .41 ({\scriptsize .35, .47}) &  .58 ({\scriptsize .49, .68}) &  .76 ({\scriptsize .69, .84}) &  .38 ({\scriptsize .30, .47}) &  .48 ({\scriptsize .39, .57}) \\
        Zhou-instruction &  .38 ({\scriptsize .30, .46}) &  .43 ({\scriptsize .35, .51}) &  .64 ({\scriptsize .54, .73}) &  .79 ({\scriptsize .72, .86}) &  .33 ({\scriptsize .26, .40}) &  .49 ({\scriptsize .41, .58}) \\
        \bottomrule
        \end{tabular}
    }
        \caption{Krippendorff’s alpha ($\alpha$) performance of prompts per model averaged over datasets.}
\label{your-label}
\end{table*}

\begin{table*}
    \centering
    \resizebox{\textwidth}{!}{%
        \begin{tabular}{lcccccc}
        \toprule
        model &                    Command-XL &                   Flan-T5-XXL &                 GPT-3.5-turbo &                         GPT-4 &                   Davinci-002 &                   Davinci-003 \\
        prompt           &                               &                               &                               &                               &                               &                               \\
        \midrule
        Direct           &  .47 ({\scriptsize .40, .54}) &  .62 ({\scriptsize .55, .69}) &  .75 ({\scriptsize .69, .81}) &  .81 ({\scriptsize .76, .87}) &  .59 ({\scriptsize .52, .66}) &  .61 ({\scriptsize .54, .68}) \\
        Kojima           &  .45 ({\scriptsize .38, .52}) &  .62 ({\scriptsize .55, .69}) &  .76 ({\scriptsize .70, .82}) &  .86 ({\scriptsize .81, .91}) &  .56 ({\scriptsize .49, .63}) &  .61 ({\scriptsize .54, .68}) \\
        Zhou             &  .55 ({\scriptsize .48, .62}) &  .58 ({\scriptsize .51, .65}) &  .75 ({\scriptsize .69, .81}) &  .89 ({\scriptsize .84, .93}) &  .65 ({\scriptsize .58, .71}) &  .66 ({\scriptsize .59, .73}) \\
        Plan             &  .53 ({\scriptsize .46, .60}) &  .62 ({\scriptsize .55, .69}) &  .73 ({\scriptsize .66, .79}) &  .84 ({\scriptsize .79, .89}) &  .55 ({\scriptsize .48, .62}) &  .63 ({\scriptsize .56, .70}) \\
        Articulate       &  .53 ({\scriptsize .46, .60}) &  .65 ({\scriptsize .58, .71}) &  .74 ({\scriptsize .67, .80}) &  .86 ({\scriptsize .81, .91}) &  .61 ({\scriptsize .54, .69}) &  .65 ({\scriptsize .58, .72}) \\
        Rephrase         &  .58 ({\scriptsize .51, .65}) &  .62 ({\scriptsize .55, .69}) &  .73 ({\scriptsize .66, .79}) &  .84 ({\scriptsize .79, .89}) &  .61 ({\scriptsize .54, .68}) &  .63 ({\scriptsize .56, .70}) \\
        Elaborate        &  .55 ({\scriptsize .48, .62}) &  .59 ({\scriptsize .52, .66}) &  .73 ({\scriptsize .66, .79}) &  .84 ({\scriptsize .79, .89}) &  .65 ({\scriptsize .58, .72}) &  .60 ({\scriptsize .53, .67}) \\
        Converse         &  .52 ({\scriptsize .45, .59}) &  .61 ({\scriptsize .55, .68}) &  .71 ({\scriptsize .65, .78}) &  .81 ({\scriptsize .76, .87}) &  .53 ({\scriptsize .46, .61}) &  .62 ({\scriptsize .55, .69}) \\
        Self-critique    &  .51 ({\scriptsize .43, .58}) &  .58 ({\scriptsize .51, .65}) &  .72 ({\scriptsize .65, .78}) &  .84 ({\scriptsize .79, .89}) &  .57 ({\scriptsize .50, .64}) &  .64 ({\scriptsize .57, .71}) \\
        Zhou-instruction &  .56 ({\scriptsize .49, .63}) &  .59 ({\scriptsize .52, .66}) &  .75 ({\scriptsize .69, .82}) &  .85 ({\scriptsize .80, .90}) &  .54 ({\scriptsize .47, .61}) &  .65 ({\scriptsize .58, .71}) \\
        \bottomrule
        \end{tabular}
    }
        \caption{Accuracy of prompts per model averaged over datasets.}
\label{your-label}
\end{table*}

\end{document}